\documentclass[10pt, a4paper]{article}
\usepackage{booktabs} 
\usepackage{multirow}
\usepackage{graphicx}
\usepackage[export]{adjustbox}
\usepackage{amsmath}
\usepackage{subcaption}
\usepackage{array}
\newcolumntype{C}[1]{>{\centering\arraybackslash}p{#1}}

\usepackage[final]{lrec2026} 

\title{Efficient Financial Language Understanding via 
Distillation with Synthetic Data}

\name{Wen-Fong (Xavier) Huang, Edwin Simpson} 

\address{
         School of Engineering Mathematics and Technology, University of Bristol \\         william.techlover@gmail.com, edwin.simpson@bristol.ac.uk\\}

\abstract{
Large instruction-following models are powerful but costly to deploy, particularly in finance, where labelled data are limited by confidentiality and expert annotation cost. We present an efficient framework for financial 
sentiment analysis
through 
\textbf{distillation with synthetic data}, 
transferring knowledge from a large instruction-tuned teacher to compact student models. 
The framework is designed for low-resource conditions, where a small set of real examples are collected and labelled by hand. 
The framework then clusters the examples and uses the clusters to select seeds for generating synthetic examples via structured few-shot prompting. 
Experiments show that clustering-based seed selection yields more representative synthetic data than random sampling, enabling compact models to achieve strong performance with minimal supervision. Notably, on a more complex and noisy text domain, the compact model trained on the complete synthetic–seed corpus even \textbf{outperforms the teacher model}, while remaining competitive on formal text. The framework provides a practical 
route toward resource-efficient domain adaptation in financial NLP with minimal human labelling effort.
 \\ \newline
 \Keywords{synthetic data, instruction distillation, financial NLP, low-resource learning, data selection}
}

\begin{document}

\maketitleabstract

\section{Introduction}

Financial sentiment analysis supports investment decision-making and market monitoring, yet annotated data are often scarce due to confidentiality and expert labelling costs~\citep{lopezlira2023chatgpt, yang2023fingpt, kirtac2024sentiment}.
Large language models (LLMs) such as GPT-4o~\citep{gpt4o} offer a potential solution: they exhibit strong instruction-following and reasoning abilities across classification, summarisation, and question answering~\citep{brown2020language}, allowing them to be applied to new tasks with minimal labelled data. However, their computational demands, latency, and proprietary nature hinder deployment in cost- and risk-sensitive settings~\citep{shen2023efficientllm, liu2025tamingtitans}.

A growing line of research transfers instruction-following behaviour from large models to smaller ones through distillation and synthetic supervision. Annotation-efficient pipelines such as \textit{Self-Instruct}~\citep{selfinstruct}, \textit{Alpaca}~\citep{alpaca}, and \textit{Orca}~\citep{orca} demonstrate that synthetic instruction–response pairs can replace large portions of human annotation, while alignment-oriented approaches like Zephyr~\citep{zephyrbeta} and LIMA~\citep{lima} show that compact, high-quality prompts can yield strong generalisation.  
In parallel, lightweight encoder architectures—DistilBERT~\citep{distilbert}, TinyBERT~\citep{tinybert}, and ModernBERT~\citep{modernbert}—provide efficient foundations for downstream adaptation.  
These trends motivate our central question: \emph{Can instruction-following knowledge be distilled from a large teacher into a lightweight, domain-specific student using only minimal labelled data?}

We introduce a reproducible framework for \textbf{distillation with synthetic data generation in financial NLP}. Unlike prior studies of financial sentiment analysis that rely on human-annotated corpora 
to fine-tune 
pretrained models like FinBERT~\citep{araci2019finbert,yang2020finbert} and FinGPT~\citep{yang2023fingpt}, our approach replaces manual labelling with structured synthetic expansion from minimal seed data and 
teacher–student distillation.
We introduce a novel coreset-style step \cite{sener2018active} for selecting a small set (12-105) of seed sentences 
by clustering  Sentence-BERT embeddings~\citep{sentencebert}, 
then
expand these seeds 
through structured prompting of GPT-4o (zero-, one-, and few-shot) to enhance the diversity of synthetic data.  
Compact students—DistilBERT and ModernBERT—fine-tuned on this synthetic–seed corpus achieve strong performance on two sentiment benchmarks: \textit{Financial PhraseBank} \citeplanguageresource{lr_phrasebank} and \textit{Twitter Financial News Sentiment} \citeplanguageresource{lr_twitter_finance}.  
On the noisier social-media domain, the compact student even \textbf{surpasses the GPT-4o teacher} in zero-shot mode.
%
Our contributions are:
\begin{itemize}
    \item A compact, reproducible framework that transfers instruction-following capability from an LLM 
    to lightweight encoder-based students.
    \item Strategies for clustering-based seed-selection and multi-prompt data expansion that increase performance with few labelled examples. 
    \item A systematic 
    evaluation for financial sentiment analysis, demonstrating effectiveness of our strategies across two diverse 
    text domains.  
    \item Empirical evidence that, on noisy financial text, a compact student trained on the complete synthetic–seed configuration can outperform its large teacher.
\end{itemize}





\section{Related Work}
\label{sec:related}

\paragraph{Distillation and Model Compression}
\label{sec:distillation}
LLMs such as GPT-4o~\citep{gpt4o} demonstrate strong instruction-following and reasoning capabilities across domains, yet they remain generalists that can lag behind fine-tuned, task-specific systems on domain benchmarks~\citep{kocon2023chatgpt, liang2023helm}. Their substantial computational requirements, latency, and closed-weight nature further constrain deployment in scenarios demanding efficiency, transparency, or confidentiality.  
Knowledge distillation provides a practical way to transfer 
a teacher model’s knowledge to smaller, more efficient students~\citep{distilbert, tinybert, modernbert}. Classical approaches such as DistilBERT~\citep{distilbert} and TinyBERT~\citep{tinybert} align intermediate representations and soft targets to reduce model size while retaining much of the teacher’s performance. ModernBERT~\citep{modernbert} refines the BERT architecture with rotary embeddings and other efficiency-oriented improvements, making it a strong candidate for 
student models. 

Beyond model compression, in text-level dataset distillation, DiLM~\citep{maekawa2024dilm}  employs clustering
within each training loop to select representative and diverse mini-batches for gradient alignment, which helps to stabilise optimisation.
In contrast, our framework applies semantic clustering prior to training
to identify diverse financial sentences as seeds for data augmentation. 

\paragraph{Synthetic Data Generation}

Building on the distillation methods above, 
synthetic data generation addresses data scarcity by producing instruction–response pairs through large teacher models. In this paradigm, the teacher not only provides supervision but also generates both the instruction and its corresponding response, creating high-quality synthetic datasets that reduce dependence on human annotation. Frameworks such as Self-Instruct~\citep{selfinstruct}, Alpaca~\citep{alpaca}, and Orca~\citep{orca} exemplify this approach, demonstrating that synthetic supervision can substantially lower annotation costs. More recent alignment-oriented methods—such as Zephyr~\citep{zephyrbeta} and LIMA~\citep{lima}—further improve data quality through preference optimisation and \textit{structured prompting}, where prompts follow predefined templates or roles that guide the model to generate consistent, label-aligned, and task-relevant outputs. Recent advances such as DiLM~\citep{maekawa2024dilm} extend this paradigm by performing text-level dataset distillation, where a language model is trained to generate synthetic text samples directly instead of optimising embedding representations.

However, these studies primarily distil general-purpose instruction-following models rather than developing domain-specific systems. Domain-sensitive applications such as financial sentiment analysis pose additional challenges, including specialised terminology, implicit market cues, and limited expert-labelled data~\citep{rodriguez-inserte-etal-2023-large,araci2019finbert}. To address these challenges, 
our pipeline expands 
a small set of domain-representative examples 
via structured prompting, balancing annotation efficiency with domain fidelity.

\paragraph{Seed Selection and Prompt Design}
Representativeness and diversity of supervision data are critical for generalisation in low-resource regimes. We employ embedding-based clustering to identify diverse and representative seeds for generating synthetic data,
using Sentence-BERT (SBERT)~\citep{sentencebert} to produce sentence-level embeddings that capture semantic similarity more effectively than previous approaches. 
Applying $k$-means clustering~\citep{lloyd1982least} to SBERT embeddings yields semantically diverse centroids, ensuring that generated data span the breadth of financial expressions rather than repeating frequent surface forms.
This idea is related to coreset selection, which  identifies samples that give good coverage of a larger training set \cite{sener2018active}.  
Prompt engineering also plays a decisive role in synthetic data quality. Structured templates inspired by Self-Instruct and Alpaca~\citep{selfinstruct, alpaca} provide predefined prompt formats that control how instructions, examples, and outputs are composed. These templates leverage \textit{in-context learning}~\citep{brown2020language}, where a few labelled instances are embedded directly within the prompt to guide model behaviour without parameter updates. This structure enables scalable expansion of small human-labelled sets into coherent, label-consistent corpora. In financial text, where subtle lexical and contextual variations influence sentiment, combining clustering-based seed selection with structured prompting could provide a reliable and reproducible basis for synthetic data generation under constrained supervision.

\paragraph{Research Gap}
Prior research shows that (1) instruction-following behaviour can be effectively distilled into compact models, (2) synthetic instruction–response data can substantially reduce manual labelling costs, and (3) domain adaptation requires careful data curation and representativeness. Yet these threads remain largely unconnected in financial NLP. Our work unifies them into a single,  framework—integrating clustering-based seed selection, structured synthetic expansion, and compact encoder training—to enable efficient domain adaptation with minimal human supervision.







\section{Methodology}
\label{sec:methodology}

\begin{figure*}[t]
    \centering
    \includegraphics[width=\textwidth]{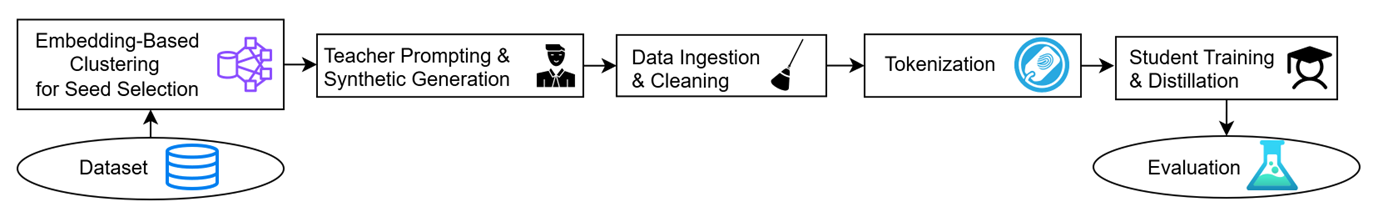}
    \caption{Overview of the proposed framework: clustering-based seed selection
    $\rightarrow$ GPT-4o prompting and synthetic generation
    $\rightarrow$ compact student fine-tuning and evaluation.}
    \label{fig:method_pipeline}
\end{figure*}

We propose a lightweight and reproducible framework for \textbf{distillation with synthetic data generation}, transferring instruction-following behaviour from a large teacher model (\textbf{GPT-4o}) to compact encoder-based students for \textbf{financial sentiment classification}.  
As shown in Figure~\ref{fig:method_pipeline}, the workflow comprises three main stages:  
(1) embedding-based clustering for representative seed selection,  
(2) structured prompting for synthetic generation, and  
(3) student model fine-tuning and evaluation.

\subsection{Embedding-Based Seed Selection}
To ensure diversity and minimise redundancy, sentences from the financial corpora are encoded using \textbf{Sentence-BERT (SBERT)}~\citep{sentencebert}.  
The variant \textit{all-MiniLM-L6-v2}~\citep{wang-etal-2021-minilmv2} produces 384-dimensional embeddings optimised for semantic similarity.  
These embeddings are clustered using $k$-means~\citep{lloyd1982least}, where the number of clusters $k$ is set to the intended number of seed examples.  
In practice, this value reflects the available annotation budget—that is, how many instances can feasibly be reviewed or labelled by domain experts.  
Each cluster therefore corresponds to one candidate seed, and the sentence nearest to the cluster centroid is selected as its representative.  

This clustering-based strategy promotes balanced semantic coverage across sentiment categories while avoiding overlap among near-duplicate samples.  
By constructing the seed set from distinct semantic regions, the framework ensures that synthetic generation originates from diverse and representative financial contexts.

\subsection{Structured Prompting for Synthetic Expansion}
Synthetic data are generated using the large instruction-following LLM \textbf{GPT-4o}~\citep{gpt4o}, accessed via the \textbf{OpenAI API}.  
The teacher model expands the clustered seed sentences into diverse, sentiment-consistent variants while preserving the intended polarity and contextual relevance.  
GPT-4o was chosen for its high reasoning capability and robust text-generation quality across both formal and informal financial language.  

Three structured prompting templates are applied to control lexical, syntactic, and contextual diversity, producing a ninefold expansion per seed after deduplication.
Each prompt template targets a different aspect of diversity, therefore, by using them in combination, we aim to ensure diverse and domain-consistent generated data. Examples of outputs from each prompt are shown in Table \ref{tab:template_examples_min}.

\paragraph{Template 1 — Few-shot, label-targeted generation.}
Defines label semantics through in-context examples before requesting a new instance expressing the same sentiment.

\noindent\fbox{%
\parbox{0.95\linewidth}{\small
\texttt{You are training a sentiment classification assistant for financial news.}\\
\texttt{Instruction: Classify the sentiment of this sentence.}\\
\texttt{Examples:}\\
\texttt{1. \{\$EX\_NEG\} → Negative}\\
\texttt{2. \{\$EX\_NEU\} → Neutral}\\
\texttt{3. \{\$EX\_POS\} → Positive}\\
\texttt{Now generate a new financial news sentence expressing: \{\$TARGET\_LABEL\}}
}}

\paragraph{Template 2 — Single-seed paraphrase expansion.}
Generates paraphrases of a labelled seed with diverse wording and structure while preserving sentiment.

\noindent\fbox{%
\parbox{0.95\linewidth}{\small
\texttt{The following sentence has sentiment \{\$LABEL\}: \{\$SEED\_SENTENCE\}}\\
\texttt{Generate 3 new sentences that:}\\
\texttt{- Express the same sentiment}\\
\texttt{- Remain realistic in the financial/news domain}\\
\texttt{- Use different wording or phrasing.}
}}

\paragraph{Template 3 — Multi-seed patterning.}
Combines multiple seeds of the same class to encourage stylistic and contextual diversity.

\noindent\fbox{%
\parbox{0.95\linewidth}{\small
\texttt{Below are examples of \{\$LABEL\} financial news:}\\
\texttt{1. \{\$SEED\_1\}}\\
\texttt{2. \{\$SEED\_2\}}\\
\texttt{3. \{\$SEED\_3\}}\\
\texttt{Generate 5 more realistic sentences with the same sentiment.}
}}

\paragraph{Design rationale.}
Template~1 establishes label semantics;  
Template~2 enhances lexical and syntactic diversity;  
Template~3 promotes intra-class generalisation.  
Together, these templates aim to produce a compact yet expressive synthetic corpus aligned with financial tone and sentiment.

\subsection{Student Model Fine-tuning}
Three compact encoder-based models—%
\textbf{DistilBERT}~\citep{distilbert}, %
\textbf{BERT-Tiny}~\footnote{\url{https://huggingface.co/prajjwal1/bert-tiny}}, and %
\textbf{ModernBERT}~\citep{modernbert}—%
are fine-tuned on the combined real and synthetic datasets using standard supervised training.  
Lower transformer layers are optionally frozen to improve stability, and early stopping is applied to mitigate overfitting.  
The trained student models can then be applied to the task of financial sentiment analysis. 




\section{Datasets and Experimental Setup}
\label{sec:datasets}

\paragraph{Datasets}
We evaluate the proposed framework on two complementary English corpora:  
(i) \textit{Financial PhraseBank}~\citep{malo2014phrasebank},\citeplanguageresource{lr_phrasebank}, consisting of formal, expert-authored statements, and  
(ii) \textit{Twitter Financial News Sentiment}~\citeplanguageresource{lr_twitter_finance}, comprising informal, real-time investor discourse.  
Labels for this second dataset are standardised to \{\textit{Negative, Neutral, Positive}\} by mapping \textit{Bearish}~$\rightarrow$~\textit{Negative} and \textit{Bullish}~$\rightarrow$~\textit{Positive}. The text data consists only of the Tweets (social media posts) themselves, without metadata such as user handles and timestamps. Examples from the two datasets are shown in the Appendix in Table \ref{tab:data_examples}.

\paragraph{Pre-processing}
Text is pre-processed to preserve sentiment-bearing cues while ensuring compatibility with each model’s tokenizer.  
Automatic cleaning steps comprised decoding escaped Unicode sequences, removing stray quotation marks and leading enumerations, and trimming extraneous whitespace.  
Aggressive normalisation steps—such as lowercasing, punctuation removal, or contraction expansion—are deliberately avoided, as stylistic features like capitalisation, repeated symbols, emojis, hashtags, and lexical emphasis often encode sentiment polarity and intensity in financial discourse.  
Stopwords are also retained to preserve syntactic and pragmatic signals (e.g., negation, modality), which are critical for accurate sentiment interpretation.

All text is tokenised using the HuggingFace tokenizer corresponding to each model.  
\textbf{DistilBERT}~\citep{distilbert} employs an uncased vocabulary, while \textbf{BERT-Tiny} and \textbf{ModernBERT}~\citep{modernbert} maintain original casing to preserve distinctions vital to financial language, including stock tickers (\texttt{\$AAPL}) and acronyms (\texttt{GDP}).  
Sequences are truncated or padded to \textbf{512 tokens} for consistency across models, aligning with their maximum supported context length.  
An \textbf{80/10/10} stratified split is applied for training, validation, and testing under fixed random seeds to ensure reproducibility and balanced label distribution.
This minimal yet model-consistent normalisation retains the linguistic richness and domain-specific stylistic variation essential for reliable sentiment representation in subsequent embedding and distillation stages.

\paragraph{Seed Sets and Synthetic Expansion}
Seed selection and prompting follow the procedures described in Section~\ref{sec:methodology}.  
Two configurations are employed: a \textbf{105-seed set}, for which we select exactly 105 representative sentences either via $k$-means clustering (one centroid-nearest sentence per cluster) or via uniform random sampling as a comparison baseline, and a \textbf{12-sample set} representing a minimal supervision baseline, also selected using the cluster centroid approach.

Each configuration is expanded using \textbf{GPT-4o}~\citep{gpt4o}, applying the three structured prompting templates introduced earlier.  
This process yields approximately a \textbf{ninefold} increase in data volume after deduplication and class balancing.  

\subsection{Training Regimes and Evaluation}
We design controlled training regimes to isolate the effects of dataset size, class balancing, layer freezing, and synthetic augmentation.
For \textit{Financial Phrasebank,}
nine regimes progressively incorporate these factors (Table~\ref{tab:pb_regimes}), serving as the main diagnostic setup. For regimes using random  selection of seed examples, each model was trained three times with different random seeds 
 to reduce the influence of single-run randomness and ensure fair comparison across models.
\begin{table}[t]
\small
\setlength{\tabcolsep}{5pt}
\renewcommand{\arraystretch}{1.05}
\begin{tabular}{@{}lp{0.80\columnwidth}@{}}
\toprule
\textbf{ID} & \textbf{Description} \\
\midrule
& \textbf{\textit{Financial Phrasebank}} \\
\midrule
(1) & Full training set (1{,}811/226/227 split) \\
(2) & 105 seeds ({\bf *} clustered); natural imbalance \\
(3) & (2) + 4 frozen layers + ES \\
(4) & (3) + balanced ({\bf **} random) \\
(5) & (3) + balanced ({\bf *} clustered) \\
(6) & (5) + Synth ({\bf **} random) \\
(7) & (5) + Synth ({\bf *} clustered) \\
(8) & 12 seeds + 4 frozen layers + ES + balanced \\
(9) & (8) + Synth \\
\bottomrule
& \textbf{\textit{Twitter Financial News Sentiment}} \\
\midrule
(10) & Full training set (9{,}544/1{,}193/1{,}194 split) \\
(11) & 105 seeds + 4 frozen layers + ES + balanced ({\bf **} random) \\
(12) & 105 seeds + 4 frozen layers + ES + balanced ({\bf *} clustered) \\
(13) & Balanced + Synth ({\bf **} random) \\
(14) & Balanced + Synth ({\bf *} clustered) \\
\bottomrule
\end{tabular}

\caption{Training regimes for sentiment classification on 
\textit{Financial PhraseBank} (IDs 1–9) and 
\textit{Twitter Financial News Sentiment} (IDs 10–14). 
Each ID corresponds to a specific experimental configuration referenced in Section~\ref{sec:results}. 
“ES” = early stopping; “Synth” = ninefold GPT-4o augmentation. 
{\bf **} = random seed selection; {\bf *} = clustered seeds (ours).}

\label{tab:pb_regimes}
\end{table}
%
For \textit{Twitter Financial News Sentiment,} five regimes evaluate robustness on shorter, noisier financial text with class distribution 65/20/15 (Table~\ref{tab:pb_regimes}), mirroring the most informative PhraseBank setups.

All regimes share identical validation and test splits to ensure comparability.  
Compact student models—\textbf{DistilBERT}~\citep{distilbert}, \textbf{BERT-Tiny}, and \textbf{ModernBERT}~\citep{modernbert}—are fine-tuned using standard optimisation settings with early stopping based on validation loss.  
Lower encoder layers are frozen in selected regimes to enhance stability and reduce computational cost.  

We evaluate the student models against the larger teacher model, \textbf{GPT-4o}, prompted directly to classify sentiment. We also compare an off-the-shelf sentiment classifier, \textbf{FinBERT}~\citep{araci2019finbert}, which combines domain-specific pretraining and  fine-tuning on gold-standard, human-labelled data from Financial Phrasebank. FinBERT uses the standard uncased BERT tokenizer (i.e., tokens are converted to lower case)\footnote{We used the classifier implementation provided at \url{https://huggingface.co/ProsusAI/finbert}}.

Performance is reported as \textbf{accuracy} and \textbf{macro-F1} on the held-out test set under a fixed random seed.






\section{Results and Analysis}
\label{sec:results}

In this section, we report the performance of the proposed synthetic data distillation framework across datasets and models, assess the contribution of each prompt template, compare clustering-based versus random seed selection, and analyse errors. 

\subsection{Overall Performance}

Table~\ref{tab:main_results} summarises the results of compact student models across datasets and training regimes.  
For regimes using random seed selection, 
we report mean results across three runs with different random initialisations.
The findings confirm that the proposed framework effectively narrows the performance gap between lightweight encoder models and the large instruction-tuned teacher (\textbf{GPT-4o}).  

\paragraph{Financial Phrasebank:}
On \textit{Financial PhraseBank} (IDs 1–9), \textbf{ModernBERT} achieves the highest score among student models, reaching \textbf{95.15\%} accuracy and \textbf{94.63\%} macro-F1 under the full synthetic–seed configuration (ID 7).  
This represents a \textbf{3.09-point gap} to the off-the-shelf FinBERT classifier (98.24/97.71) and a \textbf{2.20-point gap} to the zero-shot teacher (97.35/97.57), 
while using under 6\% of the original human-annotated data.  
\textbf{DistilBERT} performs comparably (93.83/92.76; ID 7), demonstrating that compact encoders can approximate full-scale performance when supported by high-quality synthetic supervision.

On Financial Phrasebank, FinBERT outperforms ModernBERT in the full training regime (ID 1), benefitting from its additional pretraining stage on in-domain financial text, which could be added to our approach in future work.

\paragraph{Twitter Financial News Sentiment:} On the more challenging social media dataset (IDs 10–14)—comprising shorter, noisier, and colloquial text—the advantages of synthetic instruction distillation become even more evident.  
\textbf{ModernBERT}, trained on clustered synthetic–seed data (ID 14), achieves \textbf{77.14\%} accuracy and \textbf{71.14\%} macro-F1, surpassing the GPT-4o zero-shot teacher (72.78/71.45) 
on both metrics.  
This improvement is significant ($p<0.01$) according to the McNemar's test \cite{mcnemar1947, dietterich1998} ($\chi^2=6.88$ with $p=0.0087$, see also Appendix \ref{sec:mcnemar}). 
This indicates that domain-targeted synthetic data can exceed generic large-model reasoning on unstructured financial discourse.

The off-the-shelf FinBERT classifier achieves 71.86\% accuracy and 67.39\% macro-F1 on the Twitter dataset, which is also notably lower than the student models (IDs 13 and 14). 
FinBERT was trained on Financial Phrasebank but not on the Twitter dataset, so it is to be expected that its performance is weaker on the social media dataset. 

\paragraph{Seed Selection:} Across both datasets, clustering-based seed selection consistently outperforms random sampling—particularly in low-resource regimes such as IDs (4–5) and (13–14)—yielding +3–7 F1 gains on average.  
Repeated multi-seed runs exhibit minimal variance (\textless1 F1 deviation), confirming the robustness and reproducibility of these improvements.  
Overall, structured seed selection coupled with synthetic expansion enables compact encoder models to attain strong accuracy, generalisation, and stability under minimal supervision.

\begin{table*}[!t]
\centering
\small
\begin{adjustbox}{max width=\textwidth,
                  max totalheight=\dimexpr\textheight-7\baselineskip\relax,
                  center}
\begin{tabular}{lllccl}
\toprule
Dataset & ID & Regime & Model & Accuracy (\%) & Macro-F1 (\%) \\
\midrule
\multirow{22}{*}{PhraseBank}
 & (1) & Full & DistilBERT & 96.04 & 94.95 \\
 &     &      & ModernBERT & \textbf{96.48} & \textbf{95.78} \\
 &     &      & BERT-Tiny   & 85.90 & 85.38 \\
\cmidrule{2-6}
 & (2) & 105 & DistilBERT  & 80.18 & 64.48 \\
 &     &     & ModernBERT  & \textbf{90.75} & \textbf{89.15} \\
 &     &     & BERT-Tiny    & 68.72 & 53.37 \\
\cmidrule{2-6}
 & (3) & 105 + Frz+ES & DistilBERT & \textbf{89.43} & 83.40 \\
 &     &              & ModernBERT & 88.99 & \textbf{85.79} \\
 &     &              & BERT-Tiny   & 80.62 & 79.30 \\
\cmidrule{2-6}
 & (4) & 105 + Frz+ES + Bal (Rand)$^{\dagger}$ & DistilBERT & \textbf{91.63} & \textbf{88.81} \\
 &     &                           & ModernBERT & 83.99 & 80.37 \\
 &     &                           & BERT-Tiny   & 73.32 & 61.49 \\
\cmidrule{2-6}
 & (5) & 105 + Frz+ES + Bal (Clust) & DistilBERT & 92.07 & 89.73 \\
 &     &                            & ModernBERT & \textbf{92.51} & \textbf{91.25} \\
 &     &                            & BERT-Tiny   & 80.17 & 80.27 \\
\cmidrule{2-6}
 & (6) & 105 + Frz+ES + Bal + Synth (Rand)$^{\dagger}$ & DistilBERT & 92.95 & 90.63 \\
 &     &                                    & ModernBERT & \textbf{94.57} & \textbf{93.29} \\
 &     &                                    & BERT-Tiny   & 80.03 & 75.98 \\
\cmidrule{2-6}
 & (7) & 105 + Frz+ES + Bal + Synth (Clust) & DistilBERT & 93.83 & 92.76 \\
 &     &                                    & \textbf{ModernBERT} & \textbf{95.15} & \textbf{94.63} \\
 &     &                                    & BERT-Tiny   & 88.99 & 87.69 \\
\cmidrule{2-6}
 & (8) & 12 + Frz+ES + Bal & DistilBERT & \textbf{79.30} & \textbf{67.30} \\
 &     &                   & ModernBERT & 67.40 & 55.54 \\
 &     &                   & BERT-Tiny   & 58.59 & 47.78 \\
\cmidrule{2-6}
 & (9) & 12 + Frz+ES + Bal + Synth & DistilBERT & \textbf{88.99} & 85.07 \\
 &     &                           & ModernBERT & 88.55 & \textbf{86.02} \\
 &     &                           & BERT-Tiny   & 80.18 & 74.67 \\
\cmidrule{2-6}
 &      & Publicly available, pretrained classifier   & FinBERT & \textbf{98.24} & \textbf{97.71} \\
 &      & Teacher (zero-shot) & GPT-4o  & \underline{97.35} & \underline{97.57} \\
\midrule
\multirow{13}{*}{Twitter}
 & (10) & Full & DistilBERT  & 86.52 & 82.49 \\
 &      &      & ModernBERT  & \textbf{86.60} & \textbf{82.54} \\
 &      &      & BERT-Tiny    & 82.07 & 76.01 \\
\cmidrule{2-6}
 & (11) & 105 + Frz+ES + Bal (Rand)$^{\dagger}$ & DistilBERT & \textbf{64.13} & \textbf{55.30} \\
 &      &                                       & ModernBERT & 59.90 & 50.06 \\
 &      &                                       & BERT-Tiny   & 57.73 & 47.19 \\
\cmidrule{2-6}
 & (12) & 105 + Frz+ES + Bal (Clust) & DistilBERT & 68.50 & 61.64 \\
 &      &                            & ModernBERT & \textbf{74.20} & \textbf{66.14} \\
 &      &                            & BERT-Tiny   & 56.87 & 49.06 \\
\cmidrule{2-6}
 & (13) & 105 + Frz+ES + Bal + Synth (Rand)$^{\dagger}$ & DistilBERT & 70.88 & 65.40 \\
 &      &                                               & ModernBERT & \textbf{74.34} & \textbf{68.06} \\
 &      &                                               & BERT-Tiny   & 59.59 & 56.04 \\
\cmidrule{2-6}
 & (14) & 105 + Frz+ES + Bal + Synth (Clust) & DistilBERT & 74.79 & 68.22 \\
 &      &                                    & \textbf{ModernBERT} & \textbf{77.14} & \textbf{71.14} \\
 &      &                                    & BERT-Tiny   & 64.99 & 56.46 \\
\cmidrule{2-6}
 &      & Publicly available, pretrained classifier & FinBERT & 71.86 & 67.39 \\
 &      & Teacher ( zero-shot) & GPT-4o & \textbf{\underline{72.78}} & \textbf{\underline{71.45}} \\
\bottomrule
\end{tabular}
\end{adjustbox}
\caption{Results across datasets and regimes. 
\textbf{Full} = full training set; \textbf{105} = 105 seeds (natural imbalance);
\textbf{Frz+ES} = frozen lower layers (4; Tiny=2) + early stopping;
\textbf{Bal} = class-balanced seeds (35/class);
\textbf{Synth} = +9$\times$GPT-4o augmentation;
\textbf{Rand}/\textbf{Clust} = random vs.\ clustered seeds.
$^{\dagger}$ = average over 3 random runs.}
\label{tab:main_results}
\end{table*}

\begin{figure*}[t]
  \centering
  \setlength{\tabcolsep}{4pt}
  \begin{tabular*}{\textwidth}{@{\extracolsep{\fill}} c c c}
    \subcaptionbox{Negative (Clustered seeds)\label{fig:tsne-neg-clustered}}{%
      \includegraphics[width=.315\textwidth]{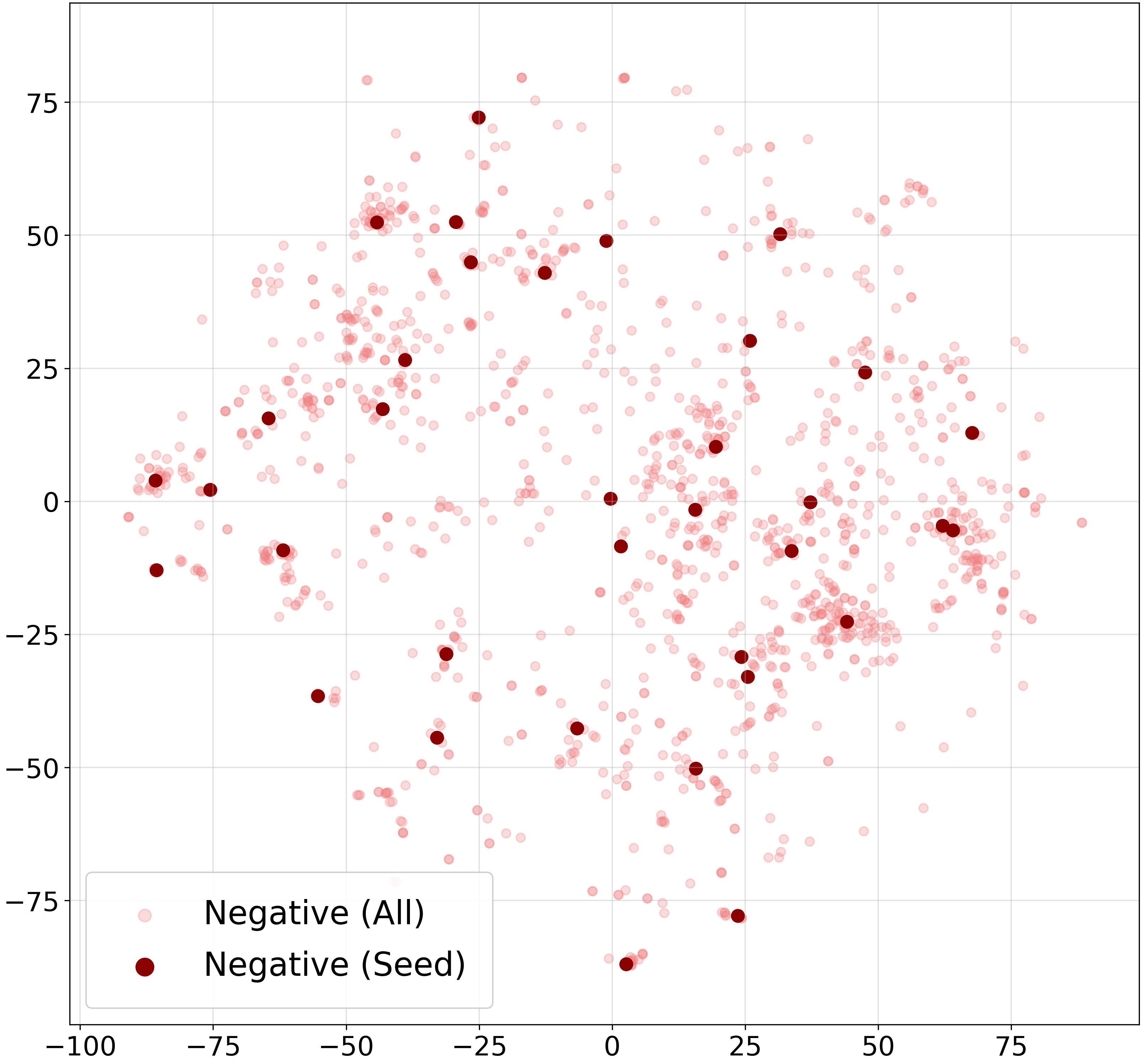}} &
    \subcaptionbox{Neutral (Clustered seeds)\label{fig:tsne-neutral-clustered}}{%
      \includegraphics[width=.315\textwidth]{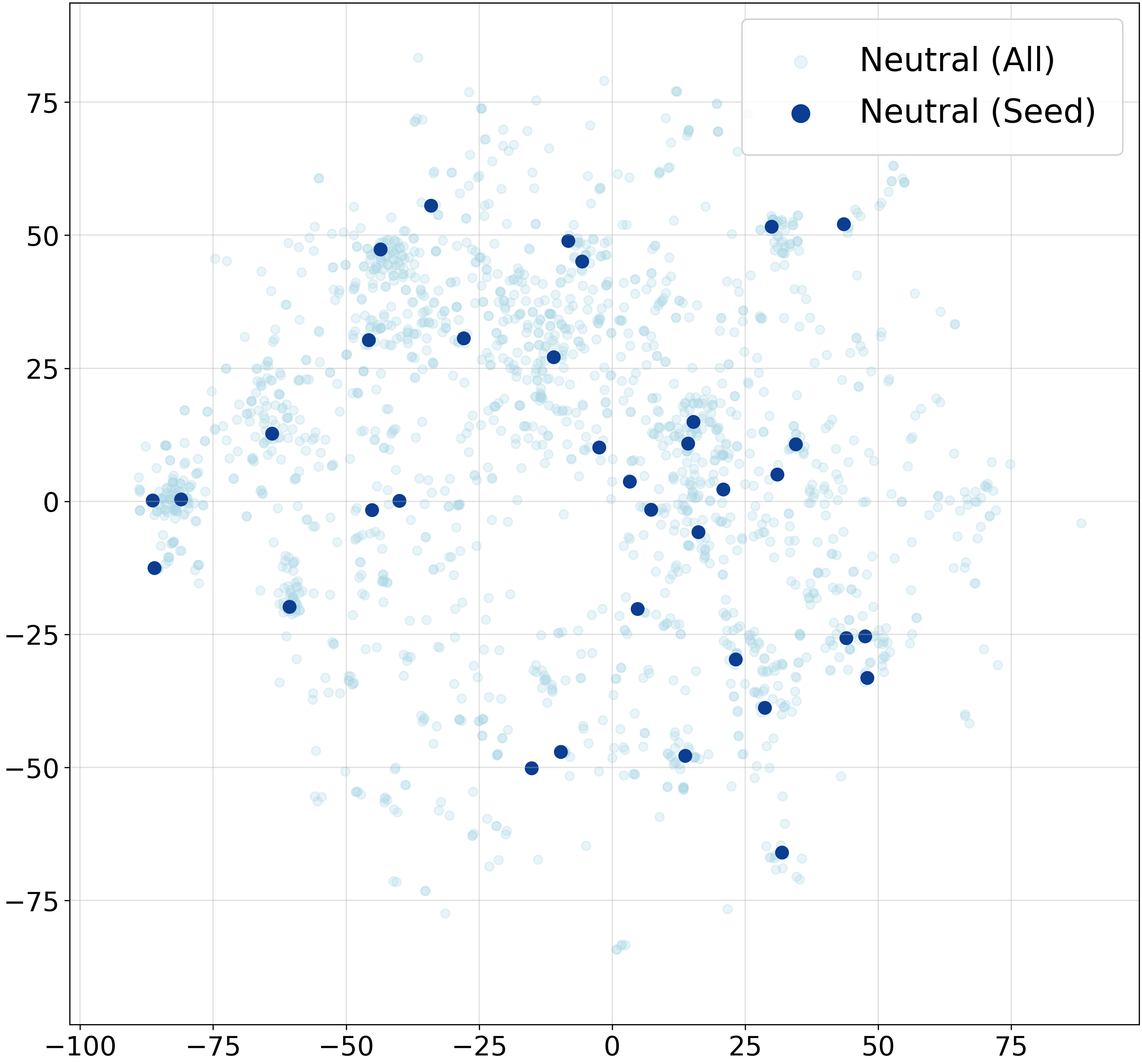}} &
    \subcaptionbox{Positive (Clustered seeds)\label{fig:tsne-pos-clustered}}{%
      \includegraphics[width=.315\textwidth]{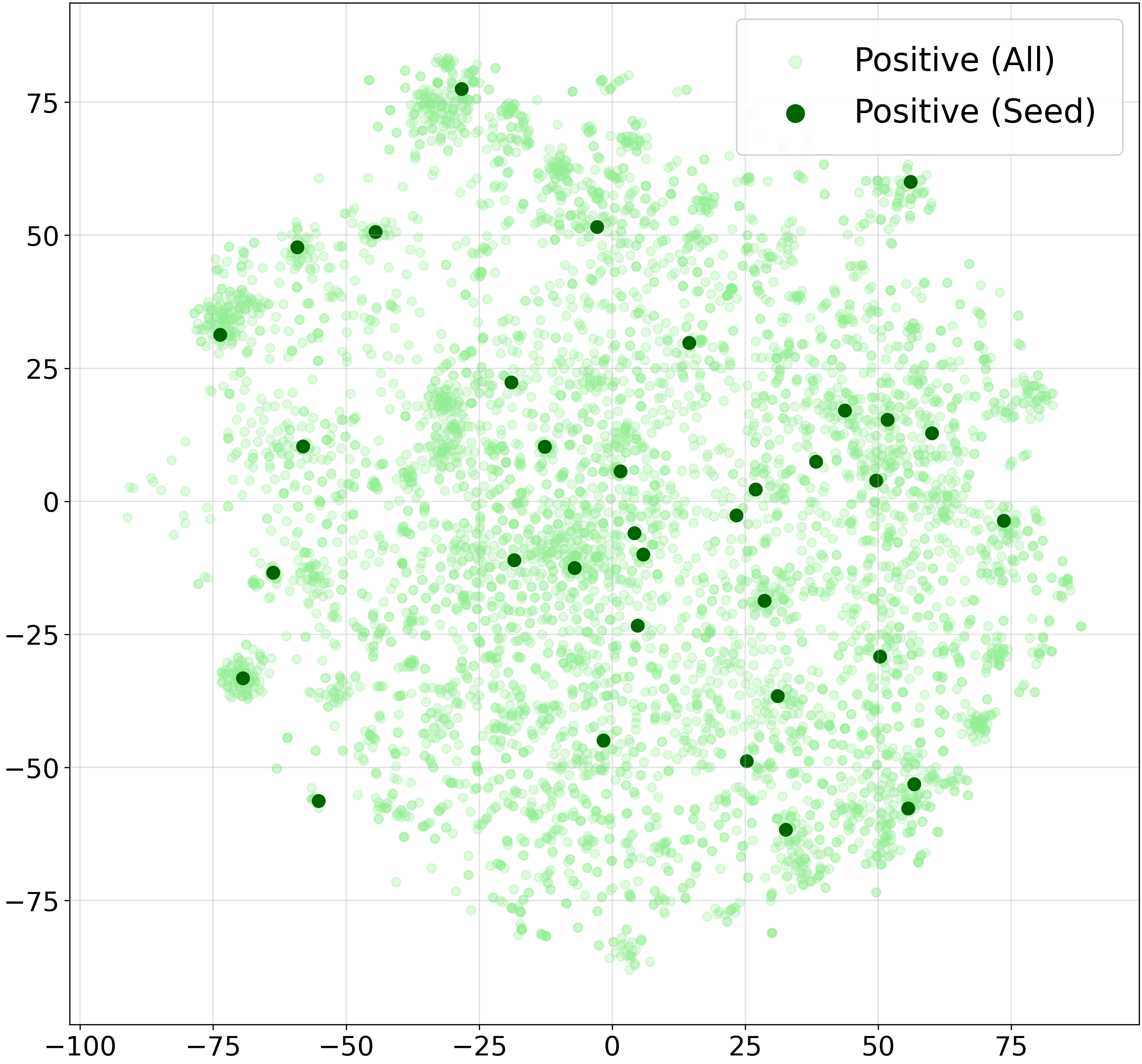}} \\[0.6em]

    \subcaptionbox{Negative (Random seeds)\label{fig:tsne-neg-random}}{%
      \includegraphics[width=.315\textwidth]{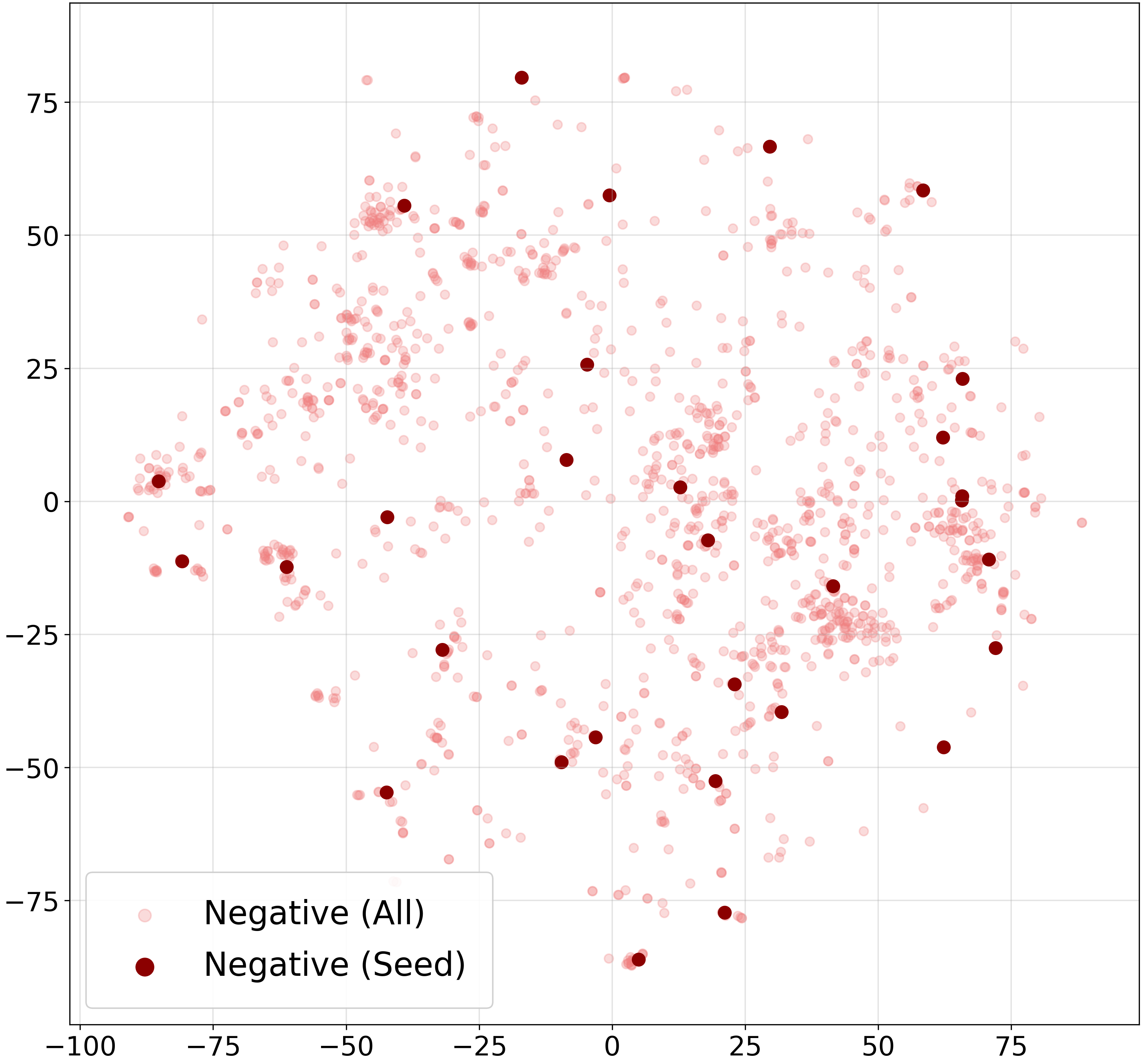}} &
    \subcaptionbox{Neutral (Random seeds)\label{fig:tsne-neutral-random}}{%
      \includegraphics[width=.315\textwidth]{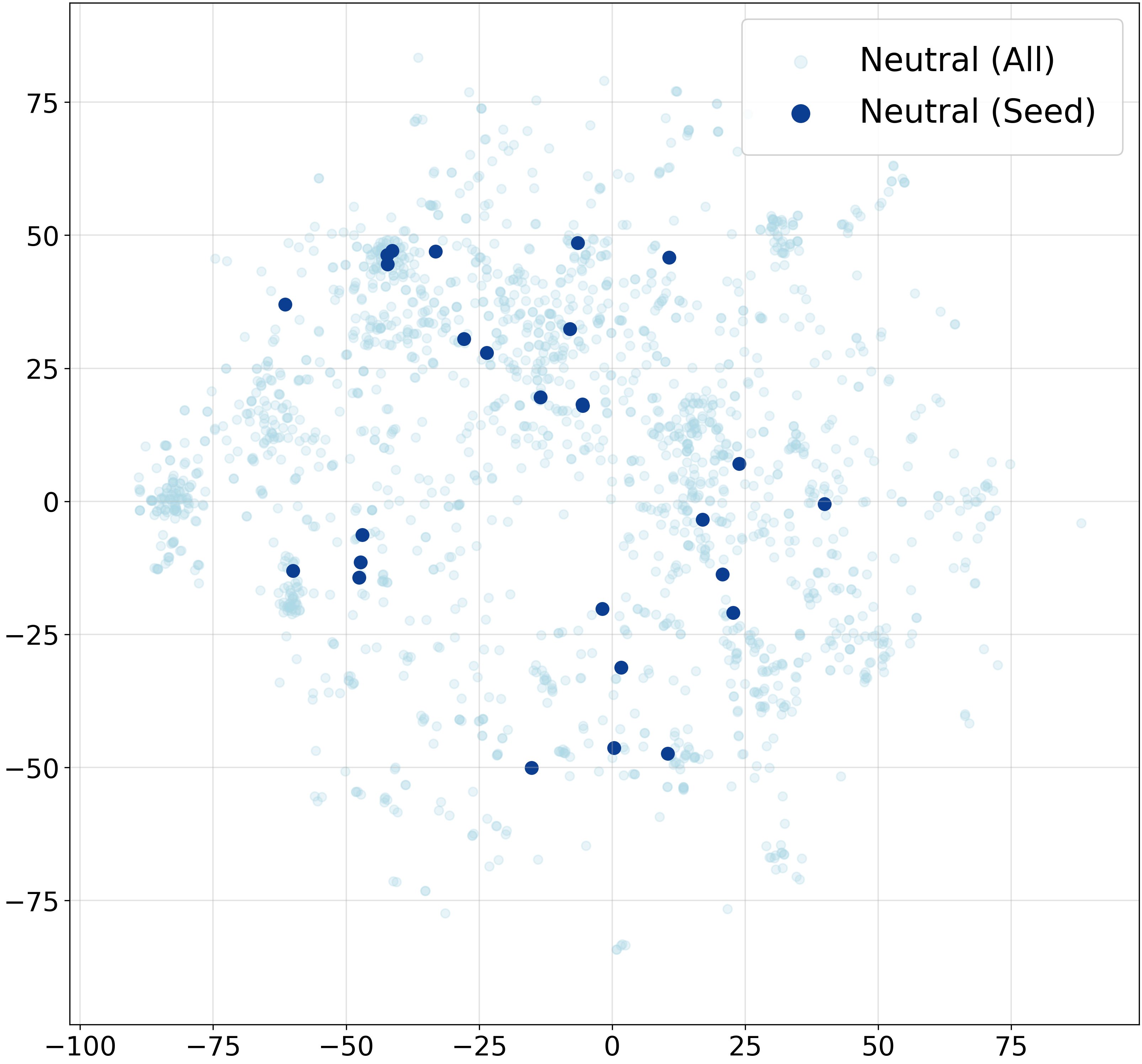}} &
    \subcaptionbox{Positive (Random seeds)\label{fig:tsne-pos-random}}{%
      \includegraphics[width=.315\textwidth]{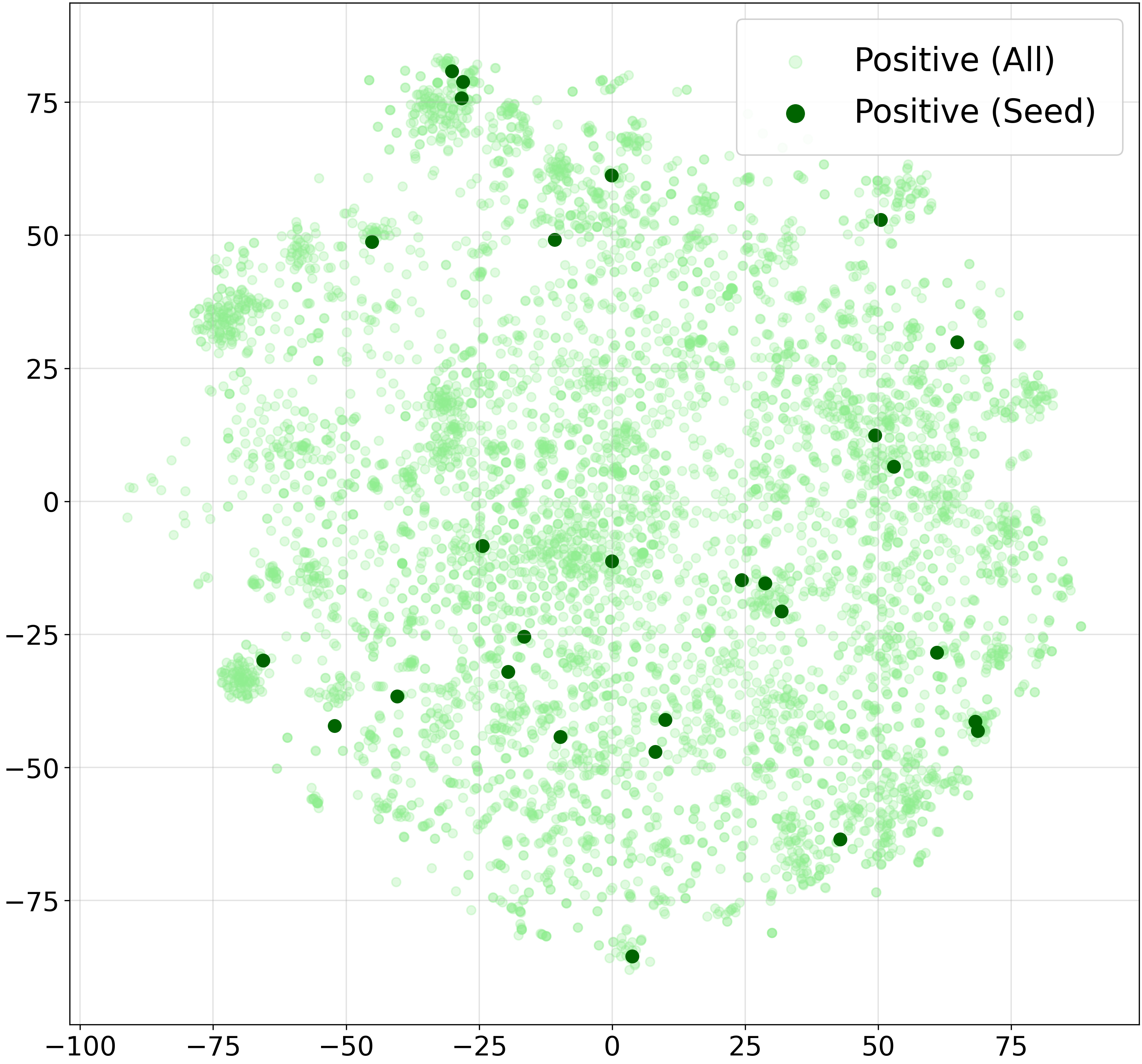}}
  \end{tabular*}
  \caption{\textbf{t-SNE comparison of seed selection strategies.}
  Columns correspond to sentiment classes (Negative, Neutral, Positive).  
  Top row: seeds chosen by \emph{K-means clustering};  
  Bottom row: seeds obtained by \emph{random sampling}.  
  All panels share a fixed t-SNE projection of PhraseBank embeddings, allowing direct spatial comparison between selection strategies.}
  \label{fig:tsne-seed-fixed-columns}
\end{figure*}

\subsection{Examples of Synthetic Data} 
\label{sec:template_examples}

Table~\ref{tab:template_examples_min} presents three representative real--synthetic pairs 
to demonstrate the complementary mechanisms of the proposed prompting templates (P1--P3), 
each targeting a particular sentiment class.  
In practice, each template is applied to generate data from all of the sentiment classes; we show one class per template here for illustration purposes only. 
The examples show how each template leverages sentiment cues from seed data to produce 
synthetic text that is lexically diverse yet sentiment-aligned.

\textbf{P1 (Neutral)} adopts a contrastive design, combining one Bearish, one Neutral, and one Bullish 
seed in a single prompt. The model observes lexical polarity cues such as \textit{“sues”} (negative), 
\textit{“moves”} (neutral), and \textit{“investments”} (positive), balancing them to generate a neutral, 
fact-focused headline. This design encourages the model to learn relational sentiment boundaries.

\textbf{P2 (Bullish)} relies on a single seed and asks for alternative rephrasings of the same sentiment. 
Here, the model retains domain-specific context (\textit{oil production}, \textit{market movement}) while 
varying syntax and word choice, demonstrating stable polarity control and lexical flexibility.

\textbf{P3 (Bullish)} provides multiple same-label seeds, prompting generalisation across related 
contexts. By abstracting common optimistic features such as \textit{price surges}, \textit{upgrades}, 
and \textit{investor confidence}, the model produces coherent yet varied bullish statements that remain 
faithful to the domain.

Together, these examples highlight how structured prompting—through cross-sentiment contrast (P1), 
within-sentiment rephrasing (P2), and multi-seed generalization (P3)—enhances realism and diversity 
in the synthetic financial corpus.

\begin{table*}[t]
\centering
\small
\setlength{\tabcolsep}{3.5pt}
\renewcommand{\arraystretch}{1.05}
\begin{tabular}{p{0.96\textwidth}}
\toprule
\textbf{Template 1 (P1 – For Neutral)}\\
\addlinespace[0.3em]
\textbf{Real Seed: }\par
\hspace{0.5em}\textit{1. Bearish: UPDATE~1--California sues e-cigarette maker Juul for selling nicotine products to youth.}\par
\hspace{0.5em}\textit{2. Neutral: Stocks making the biggest moves midday: TD~Ameritrade, Tiffany, Uber, Hasbro \& more.}\par
\hspace{0.5em}\textit{3. Bullish: \$AMZN --- Amazon: One Of The Best Long-Term Investments In The Tech Sector.}\\[0.35em]
\textbf{Synth.~e.g.: } \textit{Wall Street opens flat as investors await key economic data releases later this week.}\\
\midrule
\textbf{Template 2 (P2 – For Bullish)}\\
\addlinespace[0.3em]
\textbf{Real Seed: } \textit{Oil boosted by renewed hopes for global production cut \url{https://t.co/4tAO1U31nz}}\\[0.35em]
\textbf{Synth.~e.g.: } \textit{Hopes of a coordinated decrease in global oil production drive prices upward.}\\
\midrule
\textbf{Template 3 (P3 – For Bullish)}\\
\addlinespace[0.3em]
\textbf{Real Seed: }\par
\hspace{0.5em}\textit{1. \$AMZN -- Amazon: One Of The Best Long-Term Investments In The Tech Sector.}\par
\hspace{0.5em}\textit{2. STOCKS SURGE INTO THE CLOSE: Dow up 7.59\%, Nasdaq up 7.35\%, S\&P up 6.95\%.}\par
\hspace{0.5em}\textit{3. UnitedHealth stock price target raised to \$335 from \$310 at SunTrust RH.}\\[0.35em]
\textbf{Synth.~e.g.: } \textit{Apple Inc.\ (\$AAPL) hits a new all-time high as analysts increase price target to \$225, citing strong iPhone sales and services growth.}\\
\bottomrule
\end{tabular}
\caption{Representative real and synthetic examples grouped by prompt template. 
Each template example is shown for one sentiment type: this is for the purpose of illustrating all sentiment classes; in practice, each template is used to generate examples of all classes. 
P1 and P3 use three real seeds to guide synthetic generation, while P2 uses a single seed.}
\label{tab:template_examples_min}
\end{table*}

\subsection{Ablation Study on Prompt Templates}
\label{sec:ablation-prompts}

To evaluate the contribution of each prompting strategy, we performed an ablation study on  \textit{Twitter Financial News Sentiment}.  
All models were trained under identical conditions with 105 real seeds and 420 synthetic samples, using fixed validation and test splits.  
Each variant excludes one template from the synthetic pool to isolate its effect.

\begin{table}[t]
\centering
\small
\setlength{\tabcolsep}{5pt}
\renewcommand{\arraystretch}{1.05}
\begin{tabular}{lcc}
\toprule
\textbf{Condition} & \textbf{Accuracy (\%)} & \textbf{Macro-F1 (\%)} \\
\midrule
Full model   & \textbf{74.43} & \textbf{67.44} \\
Without P1   & 72.92 & 67.06 \\
Without P2   & 72.81 & 67.20 \\
Without P3   & 71.25 & 65.96 \\
\bottomrule
\end{tabular}
\caption{Ablation on prompt templates for the \textit{Twitter Financial News Sentiment} dataset using 105 real seeds and 420 synthetic samples.}
\label{tab:ablation_prompts}
\end{table}

As shown in Table~\ref{tab:ablation_prompts}, removing any single template reduces accuracy and macro-F1, confirming that all three templates contribute complementary diversity to the synthetic corpus.  
The largest drop occurs when Template~3 is removed, indicating that its multi-seed patterning is especially important for broadening contextual coverage and improving generalisation on Twitter text. 
Nevertheless, the relatively balanced performance across the three ablations demonstrates that each template plays a distinct, non-redundant role,  
and that combining prompts can improve on using a single template. 

\subsection{Effect of Seed Selection Strategy}
To better understand why clustered sampling improves performance, Figure~\ref{fig:tsne-seed-fixed-columns} visualises the spatial distribution of seed coverage for the \textit{Twitter Financial News Sentiment} dataset using a shared t-SNE projection of the SBERT embeddings.  
Clustered seeds (top row) show broad and evenly dispersed coverage across the embedding space, effectively representing both central and peripheral semantic regions.  
This balanced selection ensures that subsequent synthetic generation captures diverse lexical, contextual, and stylistic variations characteristic of social-media discourse, contributing to the generalisation improvements observed in Table~\ref{tab:main_results}.  

In contrast, randomly selected seeds (bottom row) tend to concentrate in high-density regions, overlooking many semantically distinct areas.  
As a result, synthetic samples generated from random seeds exhibit lower diversity and weaker domain coverage.  
Overall, both visual and quantitative evidence demonstrate that \textbf{clustering-based seed selection enhances semantic diversity and leads to more effective synthetic expansion}, particularly in noisy, low-resource domains such as financial Twitter data.

\subsection{Error Analysis Across Datasets}
To better understand model behaviour, we conducted a qualitative error analysis across both datasets using the \textbf{best-performing model, ModernBERT}. 
This analysis focuses on ModernBERT because it achieved the highest overall performance 
among all evaluated student models, making it the most representative case for examining residual classification errors and linguistic patterns.

For \textbf{Financial PhraseBank}, ModernBERT achieves consistently strong performance (95.15\% accuracy; macro-F1 = 94.63), with most errors arising from the \emph{positive} class. Specifically, 3.5\% of positive samples were predicted as negative and 7.0\% as neutral, whereas the \emph{negative} class was perfectly separated and the \emph{neutral} class exhibited minimal confusion. This trend reflects the subtle and sometimes ambiguous tone of corporate financial statements: factual phrasing can obscure positive sentiment, while comparative expressions containing explicit terms such as \emph{``negative''} may mislead the classifier. \textit{Representative misclassifications (highest-confidence examples):}

\begin{itemize}
    \item 
    \textbf{Positive $\rightarrow$ Neutral} (4/57 = 7.0\% of positive examples):  
    ``The company increased its production capacity following the completion of its new plant in Germany.''
    \emph{(confidence 91.8\%)}

    \item 
    \textbf{Positive $\rightarrow$ Negative} (2/57 = 3.5\%):  
    ``Operating profit improved to EUR~2.3~mn compared to a \emph{negative} EUR~5.1~mn a year earlier.''
    \emph{(confidence 89.4\%)}
\end{itemize}

On \textbf{Twitter Financial News}, ModernBERT reaches 77.14\% accuracy and a macro-F1 of 71.14\%. 
Misclassifications often stem from terse, context-poor posts dominated by tickers or URLs, where limited linguistic cues hinder sentiment inference. 
Among the misclassified tweets ($N{=}273$), \texttt{\$TICKER} symbols appear in 22.0\% (60/273) and URLs in 42.5\% (116/273), while emojis are absent. 
Synthetic augmentation improves robustness—particularly for \textbf{DistilBERT} and \textbf{BERT-Tiny}—yet residual errors frequently involve neutral macroeconomic or market-commentary headlines flagged as bearish, or brief bullish cues interpreted as neutral. Representative cases include:
\begin{itemize}
    \item \textbf{Neutral $\rightarrow$ Bearish}: 
    ``Did Changing Sentiment Drive Mountain Province Diamonds's (TSE:MPVD) Share Price Down A Painful 82\%?'' 
    \emph{(confidence 94.9\%)}
    \item \textbf{Bullish $\rightarrow$ Neutral}: 
    ``Stock Market Update: Netflix remains among today's winners'' 
    \emph{(confidence 95.1\%)} 
\end{itemize}

\section{Conclusion and Future Work}
\label{sec:conclusion}

This study presented a lightweight framework for \textbf{distillation from synthetic data}, enabling compact encoder models to inherit instruction-following behaviour from large teachers using a minimal number of domain-specific seeds.  
Through clustering-based seed selection and structured prompting, the framework generates semantically diverse, label-consistent data that enhances model performance while reducing annotation effort.  
With only 12–105 human-labelled sentences, distilled students achieved strong performance,
and 
ModernBERT even \emph{surpassed} the zero-shot GPT-4o teacher on noisy financial text. 
This result aligns with previous findings that show how smaller models with task-specific fine-tuning can out-perform large, generalist LLMs~\citep{kocon2023chatgpt, liang2023helm}.
Ablation results verified the importance of semantically guided seeds and combination of structured prompts, while consistent outcomes across two very different datasets demonstrated 
generalisability.

Future work could explore active learning-based seed selection~\citep{settles2009active} to improve coverage,
%
learning 
from human-annotated or model-generated explanations to enhance transparency~\citep{hase2021rationale, wiegreffe2021teach}, and calibration-aware distillation, which transfers not only predictive accuracy but also uncertainty calibration from teacher to student~\citep{muller2019calibration, karpukhin2022calibration} to 
further improve reliability.
This framework could also be extended 
beyond sentiment analysis to other finance-related tasks, such as event detection, financial risk signals, and decision reasoning.

\section{Limitations}
\label{sec:limitations}

While effective, the proposed framework exhibits several limitations.  
First, performance can vary with prompt phrasing and the representativeness of selected seeds; domain shift may reduce gains if the seed distribution fails to capture emerging financial expressions.  
Second, reliance on a single teacher model restricts linguistic variety and may propagate its biases, particularly in sentiment or stance expressions tied to market narratives.  
Third, although clustering promotes semantic diversity, it may not fully capture pragmatic or temporal nuances in evolving financial discourse.  
Finally, the framework has not yet been evaluated beyond English or in cross-domain adaptation settings.  

\noindent To ensure safe and transparent application, future studies should analyse class-wise calibration, document prompt–output dependencies, and consider incorporating human oversight or expert validation when deploying sentiment analysis models in sensitive or high-stakes financial contexts.

\section*{Data and Code Availability}
The code, trained models, and synthetic datasets used in this study are publicly available at:
\url{https://github.com/XavierHuangWF/efficient-financial-distillation}.

\clearpage

\section{Bibliographical References}\label{sec:reference}

\bibliographystyle{lrec2026-natbib}
\bibliography{lrec2026-example}

\section{Language Resource References}
\label{lr:ref}
\bibliographystylelanguageresource{lrec2026-natbib}
\bibliographylanguageresource{languageresource}

\section{Appendix}

\subsection{Dataset Examples}
Examples illustrating the text data from Financial Phrasebank and Twitter Financial News Sentiment are shown in Table \ref{tab:data_examples}.
\begin{table*}[h]
\centering
\small
\setlength{\tabcolsep}{3.5pt}
\renewcommand{\arraystretch}{1.05}
\begin{tabular}{p{0.96\textwidth}}
\toprule
\textbf{Financial Phrasebank}\\
\addlinespace[0.3em]
\textbf{Positive: }\textit{Viking Line 's cargo revenue increased by 5.4 \% to EUR 21.46 mn , and cargo volume increased by 2.4 \% to 70,116 cargo units .}\\
\textbf{Neutral: }\textit{At the request of Finnish media company Alma Media 's newspapers , research manager Jari Kaivo-oja at the Finland Futures Research Centre at the Turku School of Economics has drawn up a future scenario for Finland 's national economy by using a model developed by the University of Denver .}\\
\textbf{Negative: }\textit{Pharmaceuticals group Orion Corp reported a fall in its third-quarter earnings that were hit by larger expenditures on R\&D and marketing .}\\
\midrule
\textbf{Twitter Financial News Sentiment}\\
\addlinespace[0.3em]
\textbf{Bullish (positive): } \textit{\$BHE: Lake Street starts at Buy}\\
\textbf{Neutral: } \textit{CA\$10.60 - That's What Analysts Think Vecima Networks Inc. Is Worth After These Results}\\
\textbf{Bearish (negative): } \textit{Autodesk downgraded to underweight from neutral at JPMorgan}\\
\bottomrule
\end{tabular}
\caption{Examples data from Financial Phrasebank and Twitter Financial News Sentiment, illustrating differences in language.}
\label{tab:data_examples}
\end{table*}

\subsection{Statistical and Reproducibility Analysis}
\label{sec:mcnemar}

To assess the reliability and significance of the proposed framework, we combined statistical testing with empirical consistency analysis.  
Although each experiment was conducted under a fixed random seed, performance across datasets and configurations remained stable, indicating that the observed gains are not artefacts of random variation.  
Such consistency is particularly important for low-resource adaptation, where limited supervision can otherwise amplify stochastic effects.

Because only \textbf{ModernBERT} surpassed \textbf{GPT-4o} in raw accuracy on the Twitter dataset, we further conducted a McNemar’s test~\citep{mcnemar1947} to determine whether this improvement was statistically significant.  
McNemar’s test evaluates discordant prediction pairs—instances where one model is correct while the other is not—under the null hypothesis that both models exhibit identical error distributions.  
It is a standard non-parametric method for paired nominal data, widely used for classifier comparison~\citep{dietterich1998}.

On the 1,194 aligned Twitter test samples, ModernBERT (distilled) achieved 77.14\% accuracy compared to 72.78\% for GPT-4o.  
Among 378 discordant predictions, ModernBERT was correct in 215 cases while GPT-4o was correct in 163 cases, yielding a ratio $b/(b+c)=0.569$ (95\% CI [0.517, 0.619]).  
The continuity-corrected McNemar statistic was $\chi^2=6.88$ with $p=0.0087$, indicating statistical significance at the 0.05 level.  
This confirms that ModernBERT’s advantage on noisy financial Twitter sentiment classification is unlikely to be due to chance, but rather reflects a genuine improvement in generalisation and robustness.

\subsection{Training Hyperparameters}
\label{sec:hyperparams}
This appendix documents the experimental environment and full training configuration to ensure reproducibility.

\subsubsection{Optimisation and Training Configuration}
Across all experiments, we fine-tuned the student models using \textbf{AdamW}
with a fixed weight decay of $0.10$ and a fixed random seed (\texttt{24266}).
Unless stated otherwise, we used a training batch size of 8 and an evaluation batch size of 32.
Sequences were truncated or padded to a maximum length of 512 tokens.
A linear learning-rate schedule with warm-up was applied.
When enabled, early stopping monitored validation loss with \texttt{patience = 10}.
Table~\ref{tab:hyperparams} therefore reports only the hyperparameters that varied across model families and datasets (learning rate and the number of frozen encoder layers).

\begin{table}[t]
\centering
\small
\caption{Hyperparameter settings across datasets and models.
“Frozen” indicates the number of lower encoder layers frozen in reduced-data regimes.}
\label{tab:hyperparams}
\setlength{\tabcolsep}{4pt}
\renewcommand{\arraystretch}{1.05}
\begin{tabular}{llcc}
\toprule
Dataset & Model & LR & Frozen \\
\midrule
\multirow{3}{*}{PhraseBank}
 & DistilBERT  & $1\times10^{-4}$ & 4 \\
 & ModernBERT  & $1\times10^{-4}$ & 4 \\
 & BERT-Tiny    & $1\times10^{-3}$ & 2 \\
\midrule
\multirow{3}{*}{Twitter}
 & DistilBERT  & $1\times10^{-4}$ & 4 \\
 & ModernBERT  & $1\times10^{-4}$ & 4 \\
 & BERT-Tiny    & $1\times10^{-3}$ & 2 \\
\bottomrule
\end{tabular}
\end{table}

\subsection{Layer Freezing and Stability Notes}
DistilBERT and ModernBERT, with deeper stacks (6--12 layers) and wider hidden sizes,
achieved stable convergence with a learning rate of $1\times10^{-4}$
and could safely freeze four lower layers in reduced-data regimes.

BERT-Tiny, being shallower and narrower, required a higher learning rate
($1\times10^{-3}$) to avoid underfitting and froze only two layers,
as additional freezing removed too much trainable capacity.

On the noisier Twitter dataset, all models used stronger regularisation
(weight decay $=0.10$) and identical batch sizing (8/32)
to improve stability under short, variable-length inputs.

\subsection{Hardware and Environment}
All experiments were conducted on a ROG Strix G533QS laptop equipped with
an AMD Ryzen~9 5900HX CPU (3.30\,GHz, 8 cores), 32\,GB RAM,
and an NVIDIA GeForce RTX 3080 Laptop GPU (16\,GB VRAM).
The system ran Windows 10 Home (Version 22H2) on a 954\,GB SSD.

\end{document}